\documentclass[a4paper]{spie}  %>>> use this instead for A4 paper
\usepackage{amsmath,amsfonts,amssymb,url}
\usepackage{graphicx}
\usepackage[colorlinks=true, allcolors=blue]{hyperref}
\usepackage[caption=false]{subfig}

\title{Classification of COVID-19 cases from chest CT volumes using hybrid model of 3D CNN and 3D MLP-Mixer}

\author[a,b]{Masahiro ODA}
\author[b]{Tong ZHENG}
\author[b]{Yuichiro HAYASHI}
\author[c,d]{Yoshito OTAKE}
\author[e]{Masahiro HASHIMOTO}
\author[f]{Toshiaki AKASHI}
\author[f]{Shigeki AOKI}
\author[b,a,d]{Kensaku MORI}
\affil[a]{Information and Communications, Nagoya University, Nagoya, Japan}
\affil[b]{Graduate School of Informatics, Nagoya University, Nagoya, Japan}
\affil[c]{Graduate School of Science and Technology, Nara Institute of Science and Technology, Nara, Japan}
\affil[d]{Research Center for Medical Bigdata, National Institute of Informatics, Tokyo, Japan}
\affil[e]{Department of Radiology, Keio University School of Medicine, Tokyo, Japan}
\affil[f]{Department of Radiology, Juntendo University, Tokyo, Japan}

\authorinfo{Further author information: (Send correspondence to M. Oda)\\M. Oda: E-mail: moda@mori.m.is.nagoya-u.ac.jp, Telephone: +81 52 789 5688\\  K. Mori: E-mail: kensaku@is.nagoya-u.ac.jp, Telephone: +81 52 789 5689}

\begin{document} 
\maketitle

\begin{abstract}
This paper proposes an automated classification method of COVID-19 chest CT volumes using improved 3D MLP-Mixer.
Novel coronavirus disease 2019 (COVID-19) spreads over the world, causing a large number of infected patients and deaths.
Sudden increase in the number of COVID-19 patients causes a manpower shortage in medical institutions.
Computer-aided diagnosis (CAD) system provides quick and quantitative diagnosis results.
CAD system for COVID-19 enables efficient diagnosis workflow and contributes to reduce such manpower shortage.
In image-based diagnosis of viral pneumonia cases including COVID-19, both local and global image features are important because viral pneumonia cause many ground glass opacities and consolidations in large areas in the lung.
This paper proposes an automated classification method of chest CT volumes for COVID-19 diagnosis assistance.
MLP-Mixer is a recent method of image classification using Vision Transformer-like architecture.
It performs classification using both local and global image features.
To classify 3D CT volumes, we developed a hybrid classification model that consists of both a 3D convolutional neural network (CNN) and a 3D version of the MLP-Mixer.
Classification accuracy of the proposed method was evaluated using a dataset that contains 1205 CT volumes and obtained 79.5\% of classification accuracy.
The accuracy was higher than that of conventional 3D CNN models consists of 3D CNN layers and simple MLP layers.
\end{abstract}

\keywords{COVID-19, computer-aided diagnosis, hybrid classification model, 3D MLP-Mixer, 3D CNN}

\section{Introduction}
\label{sec:intro}  % \label{} allows reference to this section

Novel coronavirus disease 2019 (COVID-19) was recognized in December 2019.
It spreads over the world causing the large number of infected patients.
The total numbers of cases and deaths related to COVID-19 are more than 670 million and 6 million in the world by January 19, 2023 \cite{worldmeters}.
Providing appropriate treatments to a patient and prevention of infection based on diagnosis result of the patient are important.
However, because of the rapid increase of COVID-19 patients, medical institutions are suffering from a manpower shortage.
Development of a computer aided diagnosis (CAD) system for COVID-19 is pressing demanded to reduce load on medical staffs.
Reverse transcriptase polymerase chain reaction testing (RT-PCR) is used to diagnose COVID-19.
However, the sensitivity of RT-PCR is not high, ranging from 42\% to 71\% \cite{Simpson20}.
In contrast, the sensitivity of chest CT image-based COVID-19 diagnosis is reported as 97\% \cite{Ai20}.
Chest CT is effective for diagnosis of viral pneumonia including COVID-19.
CT image-base CAD systems are important in COVID-19 diagnosis.
To develop a CAD system for COVID-19, automated classification method of CT image/volume is necessary.

Many automated classification method of COVID-19 patients using chest CT images/volumes have been proposed.
Convolutional neural networks (CNN) are commonly used to classify CT images.
However, because receptive field of CNN is limited, local image features are used in classification using CNN.
Vision Transformer (ViT) \cite{vit} is becoming popular in image classification because it performs classification considering image features from whole image.
However, ViT requires a large number of training data to achieve high classification performance.
Recently, MLP-Mixer \cite{mlpmixer} is proposed as a new image classification model.
Its architecture is mainly consists of multi layer perceptrons (MLPs), which is simple and is similar to ViT.
The MLP-Mixer achieved classification performance close to ViT.
The MLP-Mixer has potential to improve classification performance of COVID-19 CAD.

We propose an automated classification method of chest CT volumes for COVID-19 CAD using a hybrid classification model that consists of both a 3D CNN and a 3D version of the MLP-Mixer.
We modified the original MLP-Mixer (for 2D image processing) to 3D version to process 3D volumes.
In our hybrid model, 3D CNN extracts local image features from CT volumes.
We apply the 3D MLP-Mixer to classify the extracted features into two classes that correspond to high or low likelihoods of COVID-19 cases.
%The 3D MLP-Mixer performs a patch-based process like ViT.
In the 3D MLP-Mixer, all local image features are mixed to perform classification based on global image features.
% image features extracted from whole 3D volume are combined to perform classification.
Hybrid use of the 3D CNN and 3D MLP-Mixer enables volume classification considering both local and global image features.

\section{Method}
\label{sec:method}

\subsection{Overview}

The proposed method classifies a chest CT volume into two classes that correspond to high or low likelihood of COVID-19 cases.
The likelihood is defined based on CT image findings confirmed by radiologists.

In the pre-processing of the classification, axial slices that do not include lung regions are removed.
Then, we apply our hybrid classification model that consists of both a 3D CNN and a 3D MLP-Mixer to the volume to perform classification.

\subsection{Pre-processing}
\label{ssec:preprocess}

From chest CT volumes, we obtain lung regions using a lung segmentation method \cite{Oda21}.
We make a sub-volume by removing axial slices that do not include lung regions from the chest CT volume.
The sub-volume is scaled to $192 \times 192 \times 128$ voxels.

\subsection{Classification using hybrid model of 3D CNN and 3D MLP-Mixer}

The sub-volume is classified into two classes that correspond to high or low likelihoods of COVID-19 cases using the hybrid classification model.
The model is trained using a training dataset and then classification is performed using a testing dataset.

\subsubsection{MLP-Mixer}

MLP-Mixer \cite{mlpmixer} for 2D image processing takes non-overlapping 2D image patches as input.
Each image patch is reshaped to a 1-dimensional (1D) vector having $C$ components.
When $S$ image patches are generated from input image, the input vector to MLP-Mixer can be represented as ${\bf X} \in \mathbb{R}^{S \times C}$.
In the MLP-Mixer, the input vector is processed by multiple layers, each layer consists of two MLP blocks.
Two MLP blocks include a cross-patch feature mixing (token-mixing) and a intra-patch feature mixing (channel-mixing) blocks.
The token-mixing and channel-mixing MLP blocks are applied to every column and row of ${\bf X}$.
The number of perceptrons in the token-mixing and channel-mixing MLP blocks are represented as $D_S$ and $D_C$, respectively.
By applying the token-mixing and channel-mixing operations, both cross-patch and intra-patch image features are extracted and utilized to perform image classification.
Output of the last layer is forwarded to a global average pooling and fully-connected layers to make final classification.

\subsubsection{Hybrid model of 3D CNN and 3D MLP-Mixer}

\begin{figure}[htb]
\begin{center}
\includegraphics[width=0.7\textwidth]{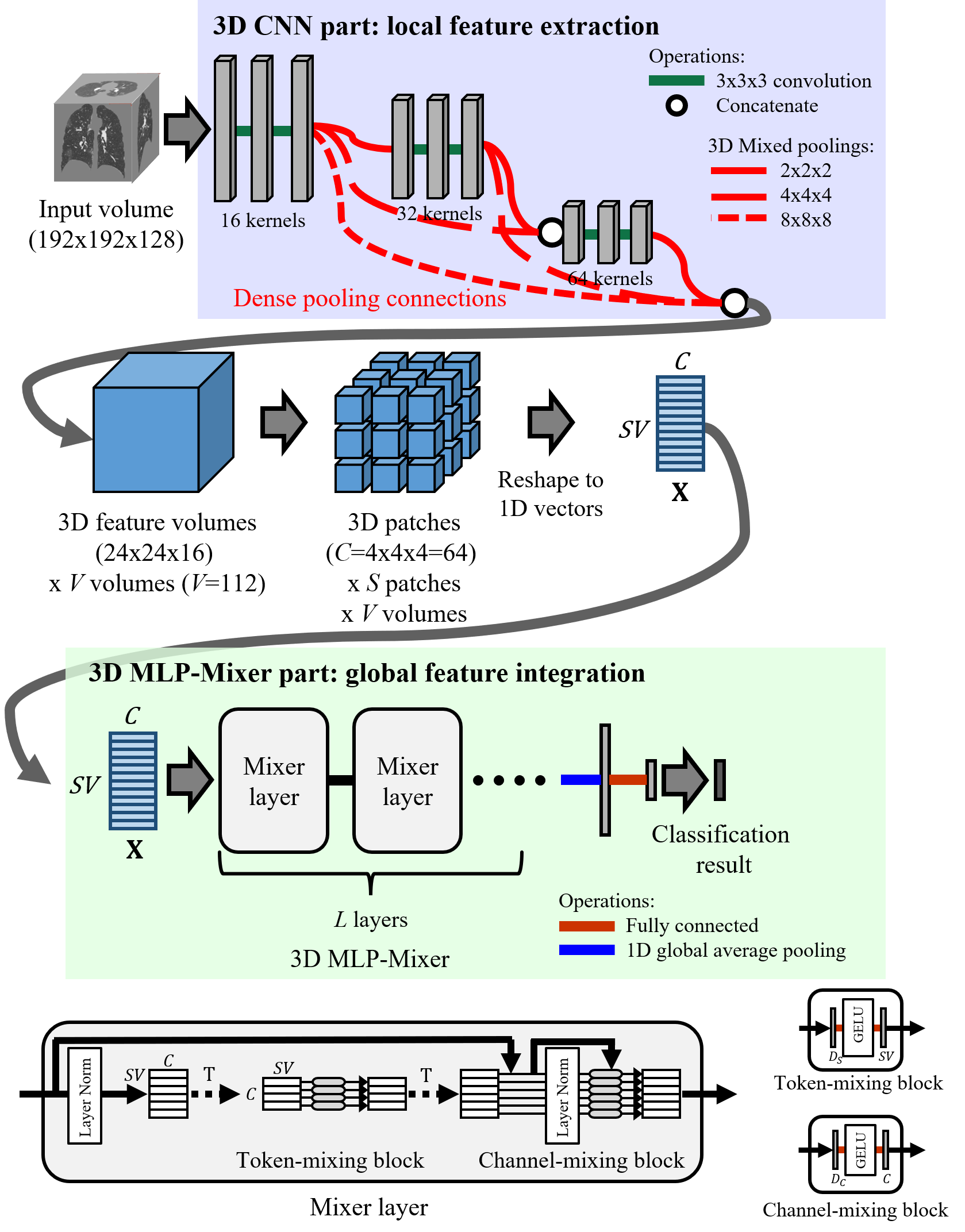}
\caption{Structure of hybrid model of 3D CNN and 3D MLP-Mixer. Input volume is processed by 3D CNN part. 3D patches are generated from feature extraction result of 3D CNN part to make input vector {\bf X} of 3D MLP-Mixer. Input vector is processed by 3D MLP-Mixer part. Finally, classification result is obtained from the last layer.}
\label{fig:processflow}
\end{center}
\end{figure}

While the MLP-Mixer has potential to perform image classification based on global image features, it requires a large number of training data to achieve high classification performance.
Commonly, a limited number of training data is available in medical image processing.
We developed a hybrid model for 3D volume classification that consists of both 3D CNN and 3D MLP-Mixer.
Its structure is shown in Fig. \ref{fig:processflow}.
The model achieves high classification performance even from a limited number of training data.
The 3D CNN part in the model performs feature extraction from a 3D volumetric image.
The 3D MLP-Mixer part performs integration of local features extracted from whole 3D volumetric image.
This combination enables classification based on volume-wide integration of local image features.

Input of the model is the scaled sub-volume generated in \ref{ssec:preprocess}.
The input is processed by the 3D CNN part that performs local feature extraction.
3D convolution operations are applied to the input data.
We use the dense pooling connections in this part.
The dense pooling connection \cite{playout18} was proposed to utilize spatial information of multiple scales in CNNs.
Mixed poolings \cite{playout18} are used in the dense pooling connections instead of max poolings to reduce information loss by max poolings.
The mixed pooling is implemented as a combination of max and average poolings.
A 3D feature volume is generated as the output of the 3D CNN part.

We make non-overlapping 3D patches from the 3D feature volume.
The size of a 3D patch is $4 \times 4 \times 4$ voxels.
Each 3D patch is reshaped to a 1D vector having $C$ components ($C=4 \times 4 \times 4=64$).
A set of all patches generated from $V$ 3D feature volumes, $S$ 3D patches are generated from each 3D feature volume, can be represented as ${\bf X} \in \mathbb{R}^{SV \times C}$.
${\bf X}$ is provided to the 3D MLP-Mixer part.

We extend the MLP-Mixer to make it applicable to 3D volumes.
We call 3D MLP-Mixer extension as the 3D MLP-Mixer.
The 3D MLP-Mixer receives a set of 1D vectors that correspond to 3D patches.
It performs classification based on the set of vectors.
In our model, the 3D MLP-Mixer receives ${\bf X}$ as the input.
It is processed by $L$ layers in the 3D MLP-Mixer similarly to the MLP-Mixer.
Output of the last layer is two class classification result.

\section{Experiments and Results}
\label{sec:results}

We applied the proposed method to 1205 chest CT volumes including COVID-19 cases and non-COVID-19 cases.
The CT volumes were obtained from multiple medical institutions in Japan.
The ground truth class labels of CT volumes were given by radiologists.
We used 80\% and 20\% of the CT volumes for training and testing, respectively.
Separations of training and testing cases were randomly performed.
We set the parameters of the proposed model as $L=2$, $C=64$, $D_S=1024$, and $D_C=256$.
In the training, minibatch size=5, training epochs=50, and Adam with learning rate=$5.0 \times 10^{-6}$ were used.

\begin{figure}[htb]
  \begin{minipage}{0.9\textwidth}
    \centering
\subfloat[]{\includegraphics[width=0.8\textwidth]{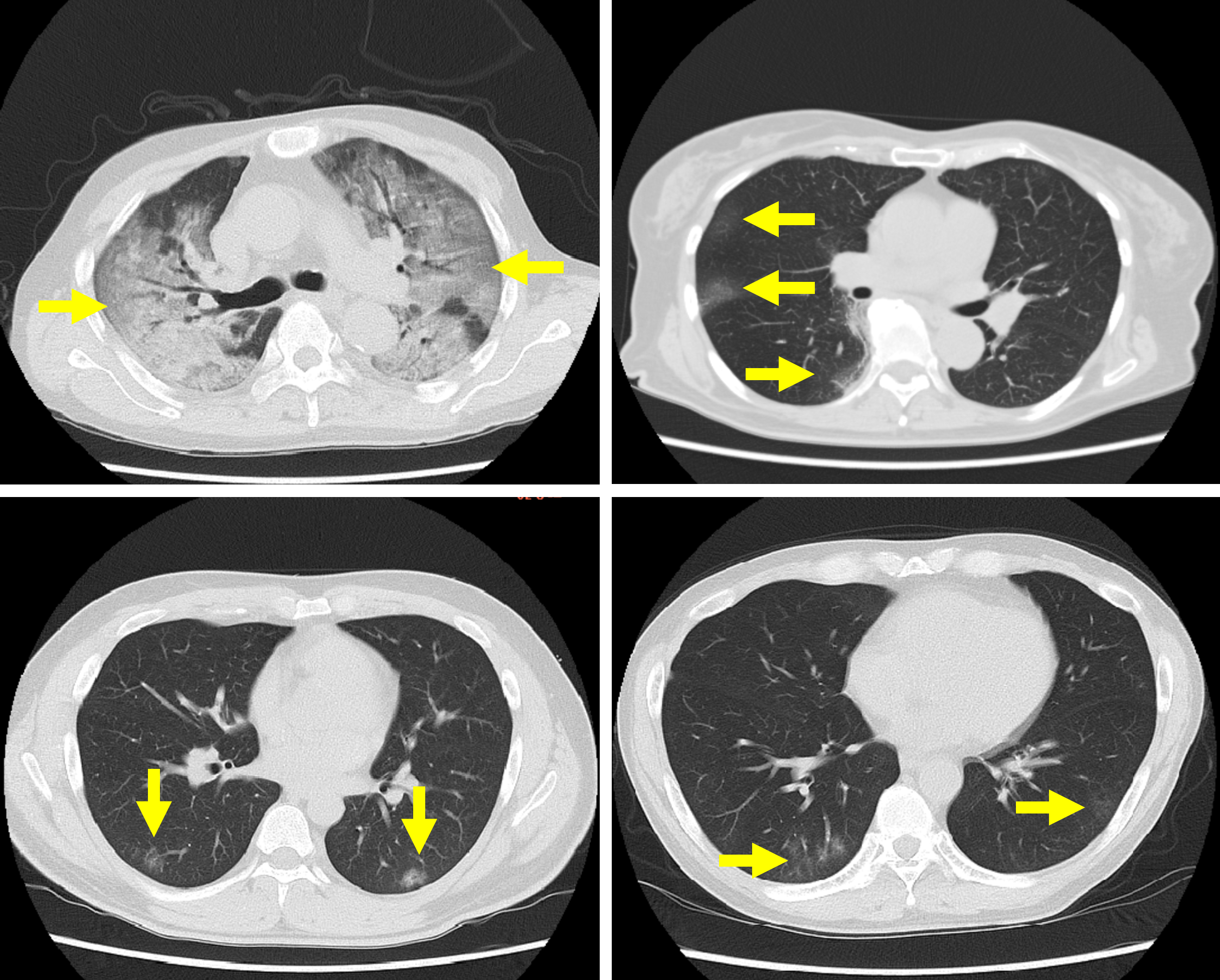}\label{fig:result_tp}}
  \end{minipage}
\\
  \begin{minipage}{0.9\textwidth}
\centering
\subfloat[]{\includegraphics[width=0.8\textwidth]{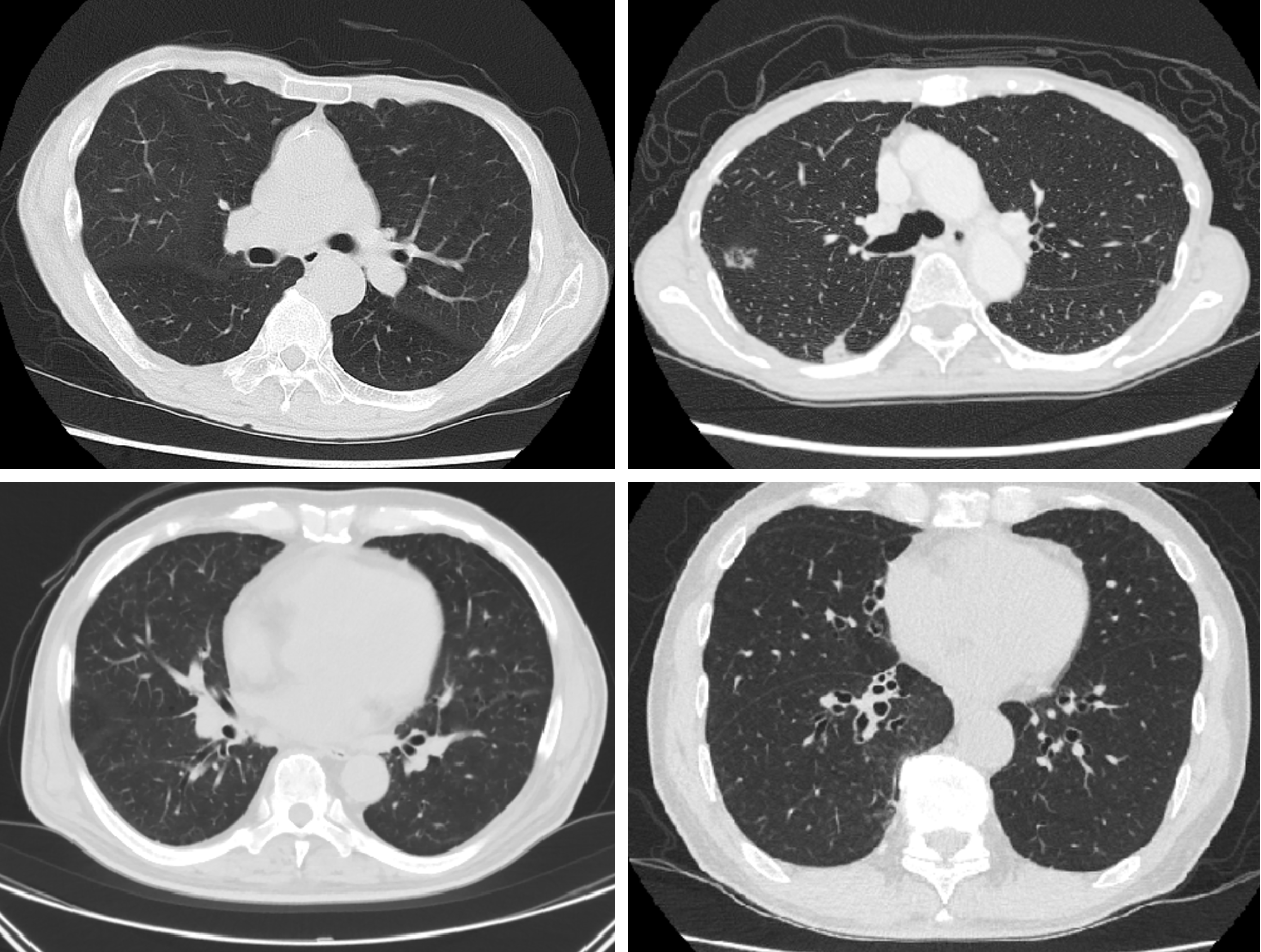}\label{fig:result_tn}}
  \end{minipage}
\caption{Axial slices of CT volumes. Infection regions are indicated by arrows. (a) cases identified as high likelihood of COVID-19 cases by radiologists. These cases were correctly classified into high likelihood class by proposed model. (b) cases identified as low likelihood of COVID-19 cases by radiologists. These cases were correctly classified into low likelihood class by proposed model.} \label{fig:result}
\end{figure}

In our experiment, classification accuracy of the proposed model was 79.5\%.
Some axial slice images of CT volumes that were correctly classified into high or low likelihoods classes by the proposed model are shown in Fig. \ref{fig:result}.
The confusion matrix of classification result is shown in Table \ref{tab:confmatrix}.

To confirm effectiveness of the 3D MLP-Mixer part in our model, we made a commonly-used 3D volume classification model that has a 3D CNN part and simple MLP layers (3D CNN model).
We trained and tested the 3D CNN model using the same dataset as the proposed model.
The classification accuracy of the 3D CNN model was 74.8\%, it was lower than the proposed model.
The result means that the 3D MLP-Mixer contributes to improve classification accuracy.

\begin{table}[htb]
\caption{Confusion matrix of classification result by proposed model. High and Low indicate high and low likelihoods of COVID-19 cases.}\label{tab:confmatrix}
\begin{center}
\begin{tabular}{|c|c|c|c|}
\hline
\multicolumn{2}{|c|}{} & \multicolumn{2}{|c|}{Estimation result}  \\ \cline{3-4}
\multicolumn{2}{|c|}{} & High & Low \\ \hline
Ground truth & High & 168 & 18 \\ \cline{2-4}
 & Low & 44 & 72 \\ \hline
\end{tabular}
\end{center}
\end{table}

\section{Discussion}
\label{sec:discussion}

The classification result of the proposed model indicated that the hybrid use of 3D CNN and 3D MLP-Mixer is effective in classification of 3D volumes.
The 3D MLP-Mixer performs classification considering global image features.
This characteristic is important in classification of viral pneumonia that causes many ground glass opacities and consolidations in large areas in the lung.
Furthermore, the hybrid model enables achievement of high classification accuracy even from a limited number of training data.
This is an advantage of the hybrid model while pure MLP-Mixer requires a large number of training data.

Further improvement of classification accuracy will be required to apply the proposed model in clinical use.
Use of data augmentation that enhances variations of anatomical shapes and intensities in training dataset, use of international dataset, and improvement of the classification model will be necessary to achieve better results.

\section{Conclusions}

We proposed a novel classification model for COVID-19 from chest CT volumes.
In image-based classification of viral pneumonia cases including COVID-19, both local and global image features are important because viral pneumonia cause many ground glass opacities and consolidations in large areas in the lung.
The proposed hybrid model of 3D CNN and 3D MLP-Mixer effectively utilizes such image features in classification.
In the evaluation using 1205 chest CT volumes, the proposed hybrid model achieved classification accuracy of 79.5\%, which was higher than the 3D CNN-based model.
Future work includes utilization of data augmentation, improvement of the number of chest CT volumes for training, and development of a COVID-19 CAD system.

\acknowledgments % equivalent to \section*{ACKNOWLEDGMENTS}       
 
Parts of this research were supported by the AMED Grant Numbers 18lk1010028s0401,
JP19lk1010036, JP20lk1010036, and 19dm0307101h0001, the NICT Grant Number 222A03, the JST CREST Grant Number JPMJCR20D5, the MEXT/JSPS KAKENHI Grant Numbers 26108006, 17H00867, 17K20099, and 21K19898, the JSPS Bilateral International Collaboration Grants. We used the Japan Medical Image Database (J-MID) created by the Japan Radiological Society with support by the AMED Grant Number JP20lk1010025.

% References
\bibliography{23spie_paper_cite} % bibliography data in report.bib
\bibliographystyle{spiebib} % makes bibtex use spiebib.bst

\end{document}